\documentclass[10pt, conference, compsocconf]{IEEEtran}

\usepackage{times}
\usepackage{epsfig}
\usepackage{graphicx}
\usepackage{amsmath}
\usepackage{amssymb}
\usepackage{subfig}

\usepackage{authblk}
\ifCLASSINFOpdf
\else
\fi
\begin{document}
%
\title{Automated Temporal Segmentation of Orofacial Assessment Videos}

\author[1,2]{Saeid Alavi Naeini}
\author[1]{Leif Simmatis}
\author[1,2]{Deniz Jafari}
\author[3]{Diego L. Guarin}
\author[1,4]{Yana Yunusova}
\author[1,2,5,6]{Babak Taati}
\affil[1]{Kite Research Institute, Toronto Rehabilitation Institute -- University Health Network}
\affil[2]{Institute of Biomedical Engineering, University of Toronto, Toronto, ON, Canada}
\affil[3]{Department of Applied Physiology and Kinesiology, University of Florida, Gainesville, Florida, USA}
\affil[4]{Department of Speech-Language Pathology, University of Toronto, Toronto, Canada}
\affil[5]{Department of Computer Science, University of Toronto, Toronto, ON, Canada}
\affil[6]{Vector Institute, Toronto, ON, Canada}










\maketitle

\begin{abstract}
Computer vision techniques can help automate or partially automate clinical examination of orofacial impairments to provide accurate and objective assessments. Towards the development of such automated systems, we evaluated two approaches to detect and temporally segment (parse) repetitions in orofacial assessment videos. Recorded videos of participants with amyotrophic lateral sclerosis (ALS) and healthy control (HC) individuals were obtained from the Toronto NeuroFace Dataset. Two approaches for repetition detection and parsing were examined: one based on engineered features from tracked facial landmarks and peak detection in the distance between the vermilion-cutaneous junction of the upper and lower lips (baseline analysis), 
and another using a pre-trained transformer-based deep learning model called RepNet (Dwibedi et al, 2020), which automatically detects periodicity, and parses periodic and semi-periodic repetitions in video data.
In experimental evaluation of two orofacial assessments tasks,
-- repeating maximum mouth opening (OPEN) and repeating the sentence “Buy Bobby a Puppy” (BBP) -- RepNet provided better parsing than the landmark-based approach, quantified by higher mean intersection-over-union (IoU) with respect to ground truth manual parsing. Automated parsing using RepNet also clearly separated HC and ALS participants based on the duration of BBP repetitions, whereas the landmark-based method could not.
\end{abstract}

\begin{IEEEkeywords}
temporal segmentation; facial movement analysis; orofacial assessment 

\end{IEEEkeywords}

%
\IEEEpeerreviewmaketitle

\begin{figure*}[t]
\centering

\subfloat[]{%
 \includegraphics[height=0.190\linewidth]{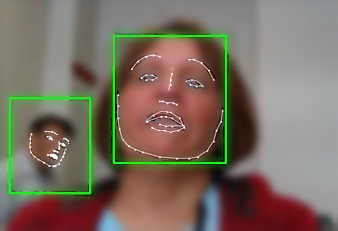}%
}\qquad
\subfloat[]{%
  \includegraphics[height=0.190\linewidth]{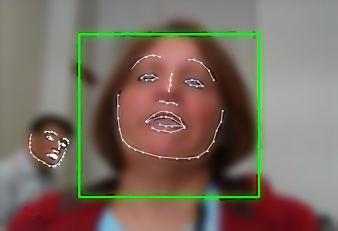}%
}\qquad
\subfloat[]{%
 \includegraphics[height=0.190\linewidth]{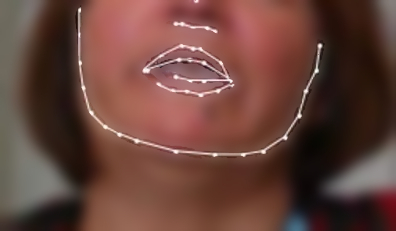}%
}

\caption{Diagram of how we obtain facial regions of Toronto Neuroface dataset in the preprocessing step; (a) detect all facial landmarks and form rectangular boxes; (b) destroy all boxes except the one closest to the camera, also add a margin to take into account small head movements; (c) the region in the corresponding box minus the eyes area. }
\label{fig_preprocessing}%
\end{figure*}




\makeatletter
\def\thickhline{%
  \noalign{\ifnum0=`}\fi\hrule \@height \thickarrayrulewidth \futurelet
   \reserved@a\@xthickhline}
\def\@xthickhline{\ifx\reserved@a\thickhline
               \vskip\doublerulesep
               \vskip-\thickarrayrulewidth
             \fi
      \ifnum0=`{\fi}}
\makeatother

\newlength{\thickarrayrulewidth}
\setlength{\thickarrayrulewidth}{4\arrayrulewidth}

\section{Introduction}
Neurological disorders such as stroke, Parkinson’s disease (PD), and amyotrophic lateral sclerosis (ALS) cause physiological deficits in the control of orofacial musculature and are linked to impairments in oro-motor abilities, speech, and emotional facial expression~\cite{Langmore, Bologna, Flowers}. A major step in diagnosing these conditions and monitoring disease progression and treatment effects is the accurate and objective assessment of orofacial impairments. 

Conventional methods to evaluate orofacial gestures and speech characteristics are perception-based, performed during examinations by clinicians, e.g., speech language pathologists.  Orofacial examinations typically involve repeating certain movements (e.g., mouth opening), syllables (e.g., /pa/, /pataka/) and sentences  multiple times. However, the reliability of these clinician-based judgements have been questioned~\cite{Kent}. 

With the strong move away from subjective clinical assessments by speech language pathologists towards objective instrumental assessments and kinematic metric extraction, automation of instrumental assessments is a high priority. There is a need to develop an accessible automated system to objectively and reliably assess orofacial gestures in non-clinical settings, allowing clinicians to identify and diagnose symptoms efficiently and accurately.

Video-based methods combined with computer vision algorithms can help improve clinical assessments using a single camera sensor~\cite{Li}. Recent studies have shown that extracting simple, yet clinically interpretable measurements (e.g., range of motion, velocity, acceleration) from jaw and lip movements can help diagnose orofacial impairments at the early stages of disease in individuals with ALS, PD, and post-stroke~\cite{Bandini01, Guarin01, Bandini02}. 

These studies involved a pre-processing step, in which a trained expert watched clinical videos and manually parsed each repetition. This process is time-intensive and requires training of professional staff~\cite{bandini05}. 
The availability of accurate models to perform temporal segmentation (parsing)  on clinical videos constitutes an important step towards automating clinical measurements, and in turn developing intelligent tools for assessing orofacial function.

Facial landmark tracking~\cite{Wu} can be used for parsing of repetitions. However, while useful on simple tasks (e.g., mouth open and close repetitions), it would be difficult to implement for the parsing of more complex tasks (e.g., repeating a sentence). In addition, even for simple tasks, implementing a landmark-based parsing algorithm requires careful engineering of subsequent signal processing steps to the specifics of the task or syllables being repeated. For instance, parsing repetitions of maximum jaw opening could be performed by peak detection in the vertical distance between the vermilion-cutaneous junction of the upper and lower lips. By contrast, parsing of another common orofacial assessment task -- pretending to smile wide with tight lips -- requires modification of the peak detection algorithm to detect distance in the horizontal direction between oral commissures (lip corners).

Therefore, it is desirable to investigate the use of data--driven parsing approaches that are less sensitive to noise and can generalize to all tasks without the need for re-engineering. The RepNet model, proposed recently by Dwibedi et al.~\cite{Dwibedi01}, is a promising technique to explore the possibility of using a deep learning model to detect and parse periodic and semi-periodic repetitions in video data. Experimental evaluation has shown that the pre-trained RepNet model generalizes well to unseen repetitions in videos in the wild~\cite{Dwibedi01}; but performance in clinical videos has not yet been validated.

In this study, we experimentally evaluate and compare the performance of RepNet and landmark-based parsing in two orofacial assessments tasks and on videos collected from individuals with ALS and healthy age-matched controls (HC). We hypothesize that RepNet-based parsing will be as good as (or better than) landmark-based parsing, as quantified by the intersection over union (IoU) between automated and manually parsed repetitions. We also evaluate automated parsing performance by comparing speech utterance duration between ALS and HC groups. Individuals with ALS typically have slower speaking rate~\cite{Masaki}, and we anticipate that RepNet-based parsing will be sufficiently accurate to preserve the separation between ALS and HC sentence utterance duration.

\begin{table*}[h]
\renewcommand{\arraystretch}{1.5}
\centering
\caption{The IoU of OPEN and BBP tasks for ALS and HC participants using Repnet and Landmark-based parsing.}
\begin{tabular}{c c rrrrrr}  
\thickhline
 &\multicolumn{2}{c}{}& \multicolumn{2}{c}{\textbf{Landmark-based}} &\multicolumn{2}{c}{\textbf{RepNet}}
\\ [0.5ex]
\cline{4-7}
&\multicolumn{2}{c}{} &\multicolumn{1}{c}{OPEN} & \multicolumn{1}{c}{BBP}&\multicolumn{1}{c}{OPEN} & \multicolumn{1}{c}{BBP}
\\ [0.5ex] 
\thickhline             
& &\textbf{ALS} &$66.7\pm1.17$ & $42.9\pm2.67$ & $70.8\pm1.81$ & $69.5\pm2.05$
\\[1ex]\raisebox{2ex}{IoU $\pm$ $95\%$ CI} & &\textbf{HC}&  $70.3\pm1.08$ & $45.3\pm1.31$ & $75.4\pm1.41$ & $70.2\pm1.71$ \\[1ex]
\thickhline                         
\end{tabular}
\label{table3}
\\ [1ex]
\end{table*}

\begin{figure*}[t]
\centering

\subfloat[]{%
 \includegraphics[width=1\linewidth]{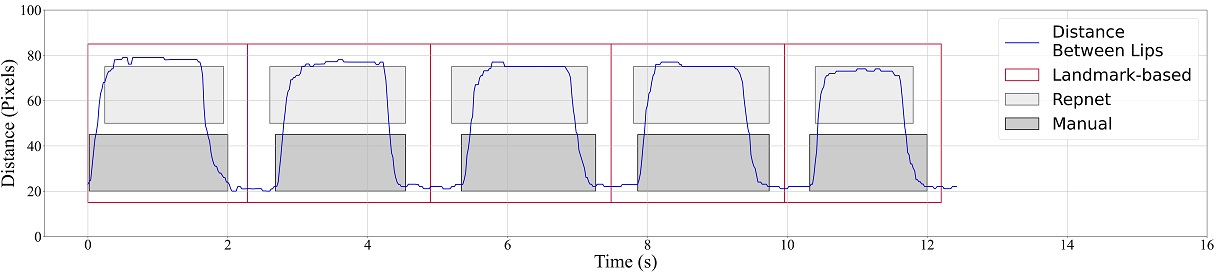}%
}\qquad
\subfloat[]{%
  \includegraphics[width=1\linewidth]{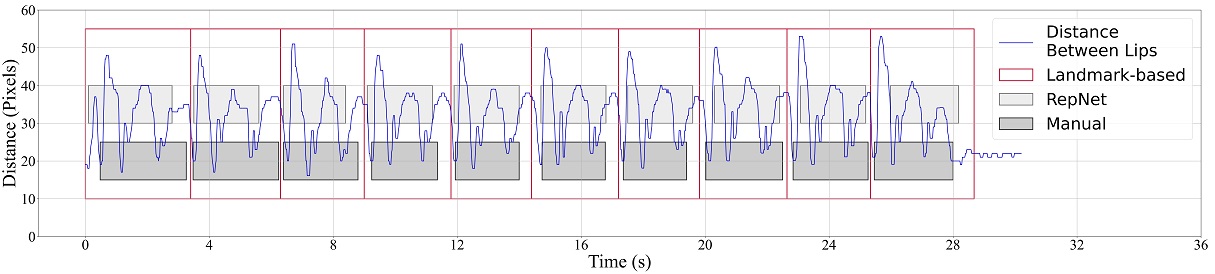}%
}

\caption{Sample visual comparison of parsing methods for (a) 5 repetitions of OPEN, and (b) 10 repetitions of BBP in an ALS participant. Height of boxes do not signify any information. }
\label{fig_VisualComparison}%
\end{figure*}

\begin{figure*}[t]
\centering

\subfloat[]{%
 \includegraphics[width=0.38\linewidth]{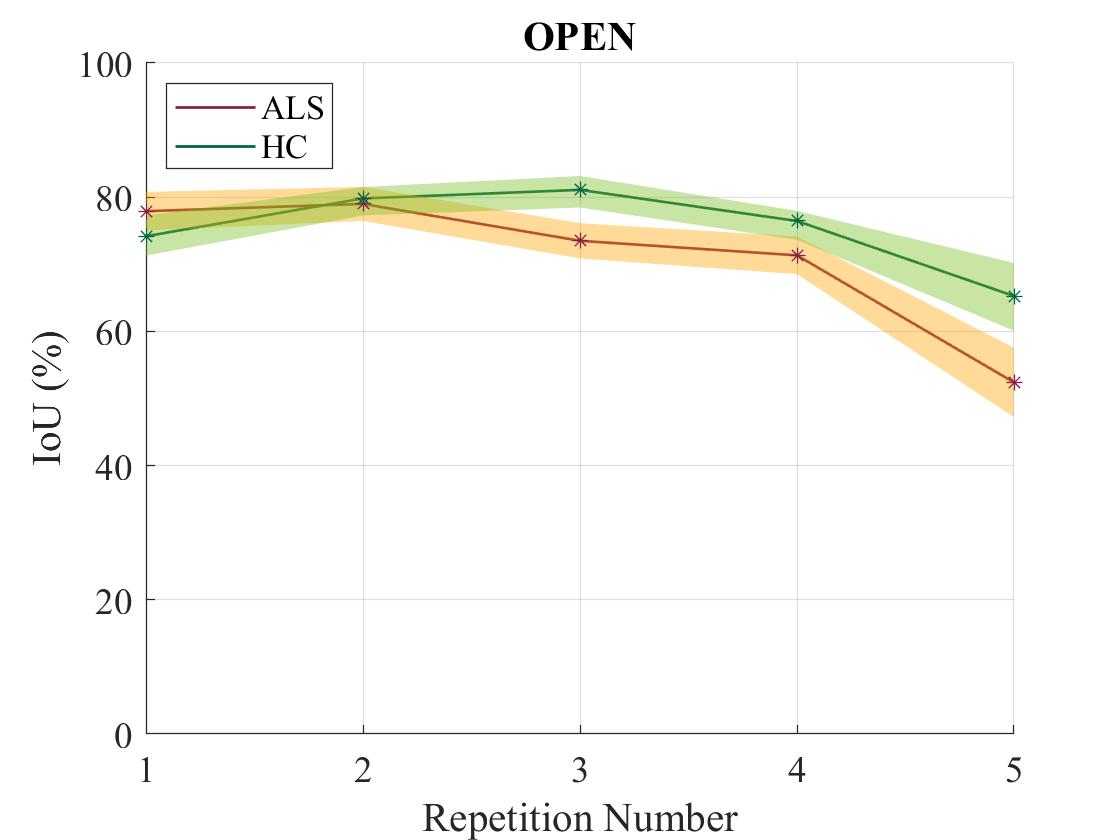}%
}\qquad
\subfloat[]{%
  \includegraphics[width=0.38\linewidth]{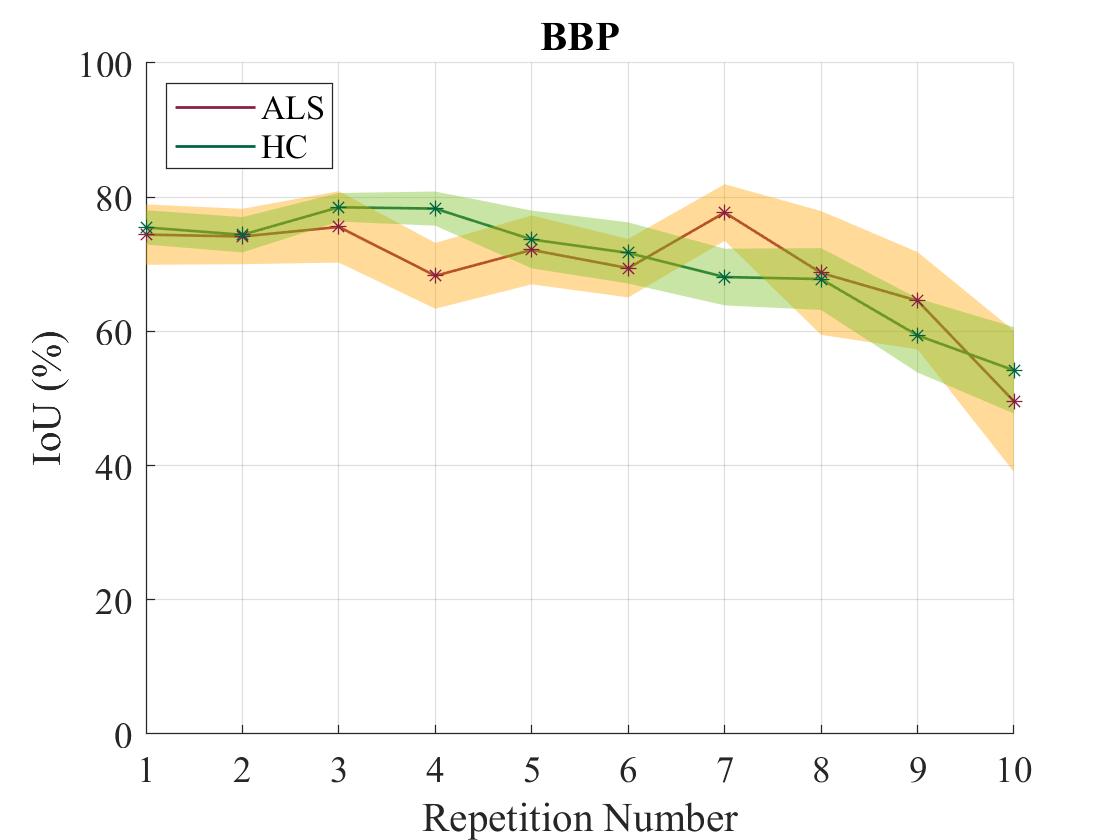}%
}

\caption{The IoU and $95\%$ confidence interval of RepNet in parsing HC and ALS participants’ performing (a) 5 repetitions of OPEN and (b) 10 repetitions of BBP.}
\label{fig_IOU}%
\end{figure*}




\section{Methods}
\subsection{Dataset}
The data used for this study was part of the Toronto NeuroFace Dataset~\cite{Bandini03}. This public dataset consists of 261 orofacial assessment videos of individuals with a neurological disorder and healthy control (HC) participants, perceptual clinical scores for each task, and manually annotated facial landmarks for a subset of the video frames~\cite{Bandini03}. In data collection step, each individual was asked to perform a set of non-speech and speech tasks frequently used by clinicians during orofacial examinations~\cite{Yunusova}. All participants were video-recorded at 50 frames per second, at 640 × 480 resolution, and filmed at face-camera distance of 30 to 60~cm. The dataset contains videos from 11 HC participants (age: $63.2\pm14.3$), 11 ALS participants (age: $61.5\pm8.0$), and 14 post-stroke (PS) participants (age: $64.7\pm14.7$)~\cite{Bandini03}. For this study, we selected the ALS and HC subset of the Toronto NeuroFace Dataset performing two tasks: maximum opening of jaw 5 times (OPEN) and repeating the sentence “Buy Bobby a Puppy” (BBP) 10 times at a comfortable speaking rate. Recorded videos were manually parsed by a trained research assistant, and these manual parsings were used as the ground truth in this study.

\subsection{RepNet}
The RepNet model first constructs a temporal self-similarity matrix (TSM) as an intermediate representation bottleneck and then applies a transformer network on the TSM to detect periodicity and estimate the period length for periodic or semi-periodic repetitions~\cite{Dwibedi01}. To construct the TSM, RepNet first applies a convolutional neural network encoder to every video frame to generate per-frame embeddings. As the RepNet model does not directly provide information about the start and end frames of each repetition, here we estimated the repetitions based on cumulative sum of per-frame periodicity scores. Even though RepNet has been trained on short videos ($\sim$10$\,$s), it can detect repetitions with longer temporal extent by choosing a higher input-frame stride~\cite{Dwibedi01}.

In preliminary testing, we noticed two common failure modes when applying the pre-trained RepNet model to videos from the Toronto NeuroFace Dataset. The first failure mode, also mentioned in the original RepNet paper, was the double counting error~\cite{Dwibedi01}. This error occurs when the model lumps each two consecutive repetitions as a single repetition and reports half the number of the counts annotated by manual parsing. The second failure mode occurred when a moving object or another person appeared in the background (moving clutter). In such cases, even after the moving clutter left the scene (e.g., the person in the background was no longer in view), it took the model a few seconds to re-adjust to repetitive motions of the mouth and jaw region of the face. To avoid these failure modes, we cropped the videos to a fixed (i.e., not moving) bounding box around the mouth and lower jaw region (see Fig.~\ref{fig_preprocessing}). This reduced the occurrence of the first error mode and eliminated the second. Cropping was done automatically, using facial landmark detection. To take (usually small) movements of the person into account, a bounding box over the union of the mouth/lower jaw region was used throughout the length of the video sequence.

Depending on the severity of neurological disorders in patients, and the duration of speech/non-speech tasks, repetitions may occur over shorter or longer temporal scales. Table~\ref{table1} (ground truth columns) presents the average length of each OPEN and BBP repetition for ALS and HC participants. As the table shows, average lengths vary from $1.28\pm0.48$ to $2.18\pm1.00$ seconds depending on the task (OPEN or BBP) and condition (HC or ALS). In order to correctly detect the number of repetitions and extract the start and end frames of each repetition, an appropriate stride should be chosen. While the trained RepNet model is not sensitive to the exact value of the stride parameter, values that are substantially (e.g., an order of magnitude) different from expected period lengths negatively impact performance. In this study, we ran the model multiple times on each video with different strides and selected the stride value that resulted in the highest estimated periodicity score.

\begin{table*}[h]
\renewcommand{\arraystretch}{1.5}
\centering
\caption{The mean durations of OPEN and BBP repetitions for ALS and HC participants. (mean $\pm$ standard deviation).}
\begin{tabular}{c c rrrrrrr}  
\thickhline
 &\multicolumn{2}{c}{}&\multicolumn{2}{c}{\textbf{Ground Truth}} & \multicolumn{2}{c}{\textbf{Landmark-based}} &\multicolumn{2}{c}{\textbf{RepNet}}
\\ [0.5ex]
\cline{4-9}
&\multicolumn{2}{c}{}&\multicolumn{1}{c}{OPEN} & \multicolumn{1}{c}{BBP} &\multicolumn{1}{c}{OPEN} & \multicolumn{1}{c}{BBP}&\multicolumn{1}{c}{OPEN} & \multicolumn{1}{c}{BBP}
\\ [0.5ex] 
\thickhline             
& &\textbf{ALS} &$1.49\pm0.81$ & $2.18\pm1.00$ & $1.79\pm0.51$ & $1.89\pm1.23$ & $1.66\pm0.81$ & $2.24\pm1.18$ 
\\[1ex]\raisebox{3ex}{Repetition Length} & &\textbf{HC}&  $1.32\pm0.42$ & $1.28\pm0.48$ & $1.50\pm0.49$ & $1.70\pm0.99$ & $1.42\pm0.47$ & $1.50\pm0.63$ \\[1ex]
\thickhline                         
\end{tabular}
\label{table1}
\end{table*}

\begin{table*}[h]
\renewcommand{\arraystretch}{1.5}
\centering
\caption{Mann-Whitney U test Parameters comparing the duration of BBP repetitions in ALS vs. HC participants. P-values smaller than 0.05 are marked with a star.}
\begin{tabular}{c c rrrrr}  
\thickhline
 &\multicolumn{2}{c}{}& \textbf{Ground Truth} & \textbf{Landmark-based} &\textbf{RepNet}
\\ [0.5ex] 
\thickhline             
& &\textbf{U-value} &$0$ & $33$ & $0$
\\[1ex]\raisebox{2ex}{Parameter Value} & &\textbf{p-value}&  $1.83\mathrm{e}{-4}^{\ast}$ & $0.21$ & $1.81\mathrm{e}{-4}^{\ast}$ \\[1ex]
\thickhline                         
\end{tabular}
\label{table2}
\\ [1ex]
\end{table*}

\subsection{Landmark-based Parsing}
We used the face alignment network (FAN)~\cite{Bulat} fine-tuned to the Toronto NeuroFace Dataset~\cite{Guarin02} for facial landmark tracking. The model detected the pixel coordinates of 68 standard facial landmarks on each face, and the vertical (y) distance between landmarks 52 and 58 was used to represent the distance between the vermilion-cutaneous junction of the upper and lower lips.

The output signal was first smoothed using 3\textsuperscript{rd} order Butterworth low-pass filter with 0.03 and 0.02 normalized critical frequencies (Nyquist~frequency~=~1) for OPEN and BBP, respectively. Finally, local minima of the smoothed signals were set as the estimated start and end frames of each repetition.

\subsection{Evaluation Criteria}
Performance was quantified in terms of the mean intersection-over-union (IoU) between detected and manually parsed repetitions. 

To further qualitatively evaluate the parsing of the videos, we also present the parsed duration of each repetition for ALS and HC participants. Individuals with ALS typically speak at lower speed~\cite{Masaki}. To investigate whether automated parsing was sufficiently accurate and precise to preserve this separation, we compared the duration of parsed BBP repetitions via the Mann-Whitney U Test across the two groups, i.e. ALS vs. HC.
\begin{figure*}[t]
\centering

\subfloat[]{%
 \includegraphics[width=0.3\linewidth]{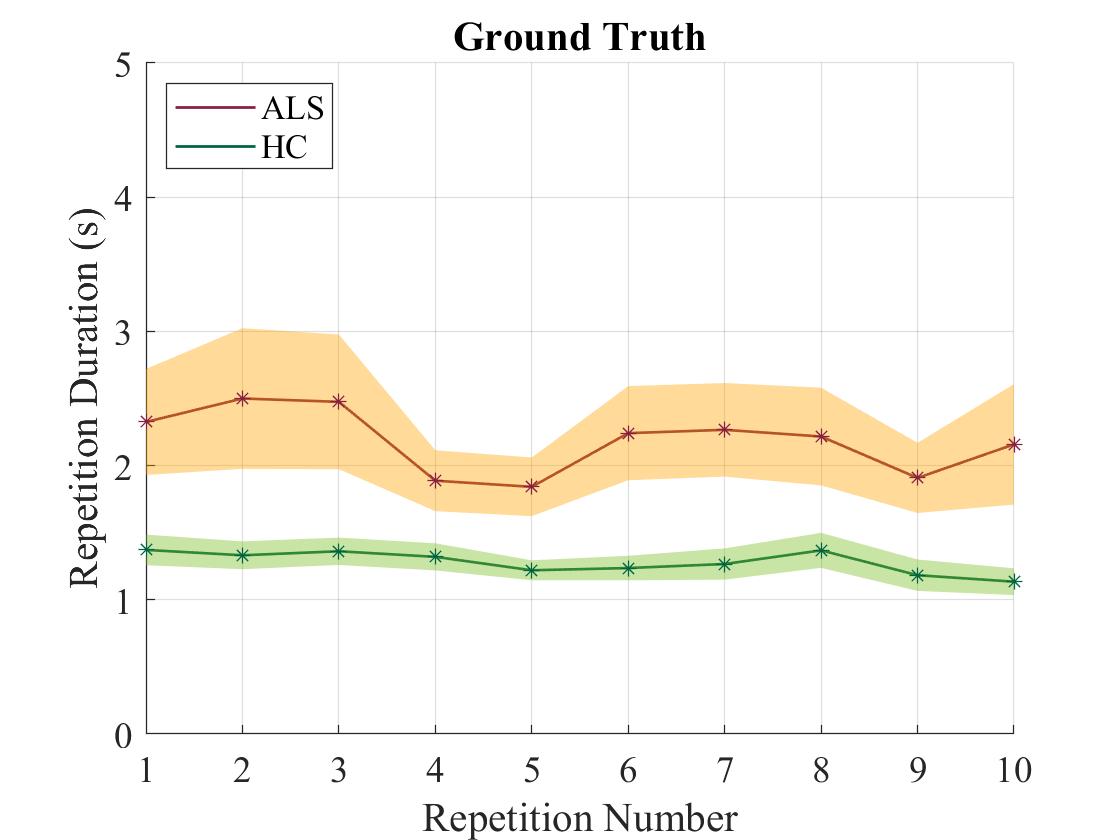}%
}\qquad
\subfloat[]{%
  \includegraphics[width=0.3\linewidth]{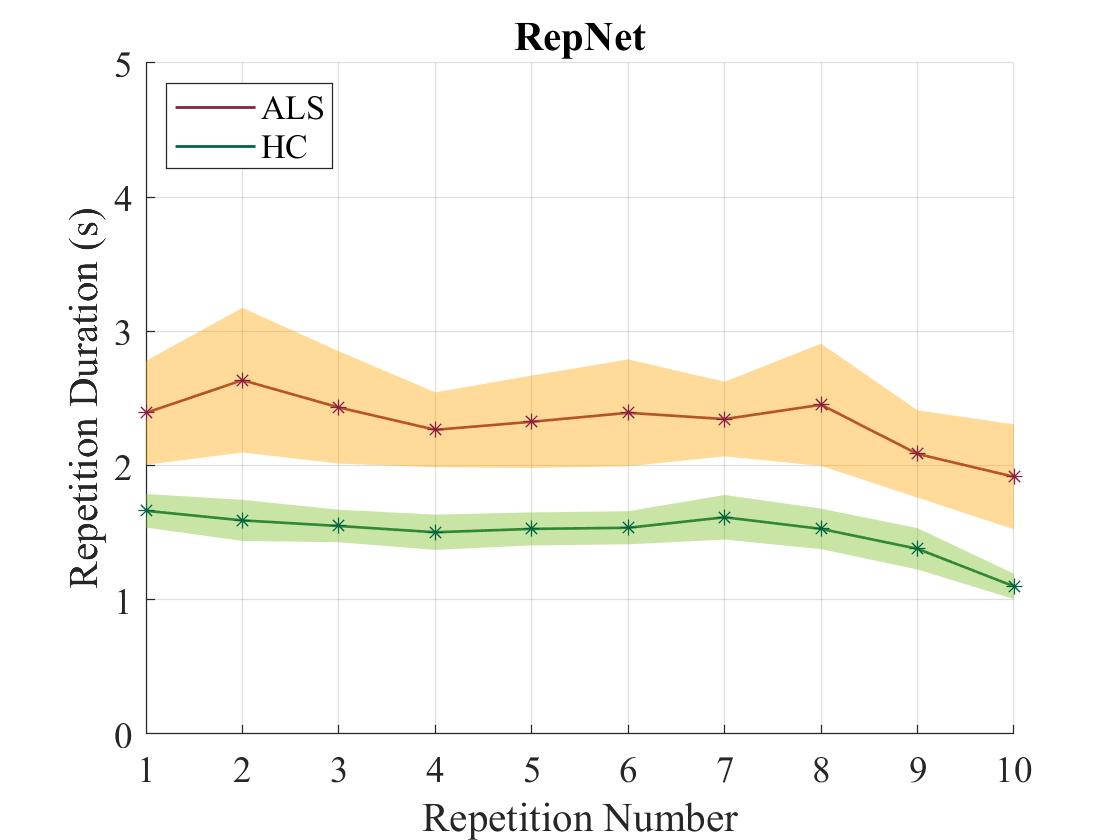}%
}\qquad
\subfloat[]{%
 \includegraphics[width=0.3\linewidth]{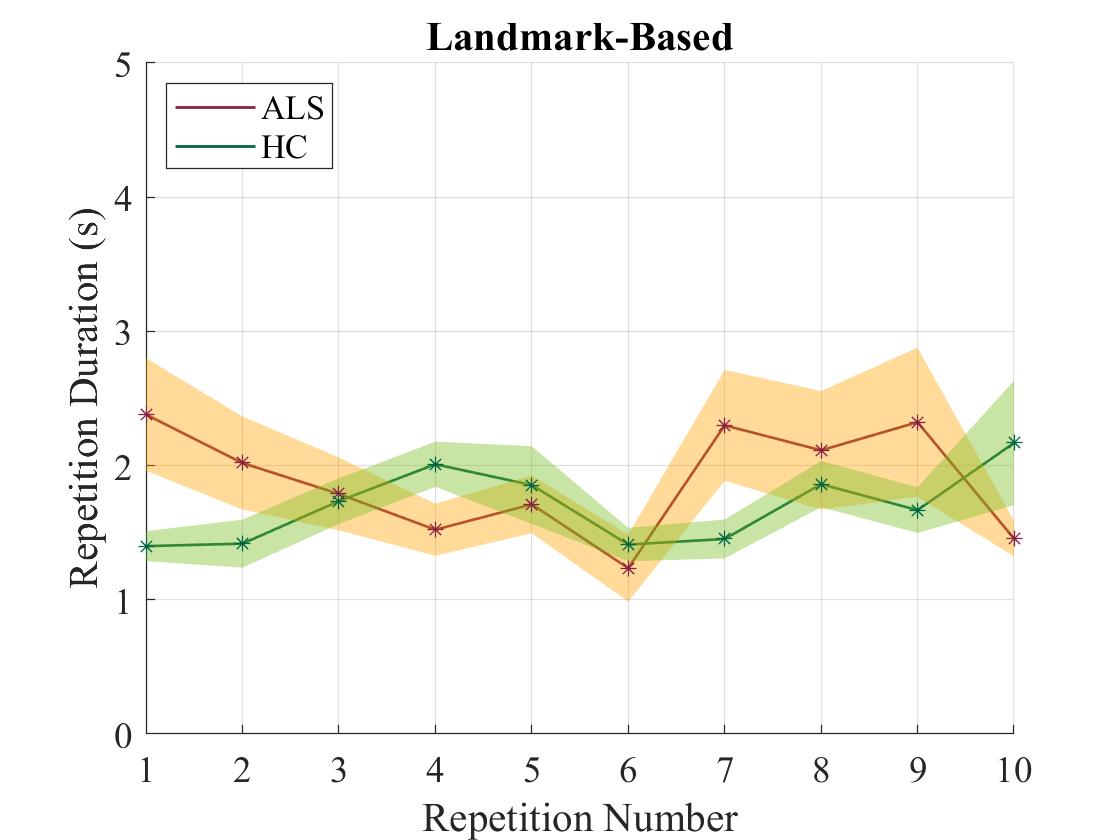}%
}
\caption{BBP repetition durations as (a) parsed manually (ground truth) or with (b) RepNet or (c) landmark-based parsing.}
\label{fig_BBP_reps}
\end{figure*}
\section{RESULTS}
Table~\ref{table3} reports the IoU of RepNet and landmark-based parsing in each task and condition. The best result in the parsing of videos was obtained in the OPEN task using RepNet. The IoU values for HC individuals were higher than those of ALS participants in both tasks and using either of the two parsing methods, perhaps due to the nature of orofacial impairments in ALS patients. While both RepNet and the landmark-based method performed well (IoU $>65\%$) in parsing of the OPEN videos, only RepNet was able to repeat this performance in BBP videos, as landmark-based parsing of BBP videos was poor (IoU $<50\%$). Even when using RepNet, a wider confidence interval (CI) and relatively higher variability were observed in the parsing of BBP repetitions (see Fig.~\ref{fig_IOU}). It was also interesting to see that IoU tended to decrease with repetition number. Performance visualization of each method with respect to manual parsing is illustrated in Fig.~\ref{fig_VisualComparison}.

Table~\ref{table1} shows the average and standard deviation of OPEN and BBP video durations in ALS and HC participants, as parsed manually (ground truth) or with RepNet or landmark-based parsing. In all cases, the true duration of each BBP and OPEN repetition was longer in ALS participants than it was in HC individuals. It was also interesting to observe that the repetition duration in the ALS group seemed to be more variable.

Table~\ref{table2} presents the result of Mann-Whitney U Test to compare durations of the BBP task in ALS vs. HC participants using manual parsing (ground truth), landmark-based parsing, and RepNet. The differences between repetition lengths were statistically significant for ground truth and RepNet, but not with repetitions detected via landmark-based parsing, meaning that RepNet model was able to preserve the separation between ALS and HC in terms of BBP repetition duration. This is further illustrated in Fig.~\ref{fig_BBP_reps}.

\section{DISCUSSION AND FUTURE WORK}

The current gold standard for the parsing of clinical videos is the manual approach which is time--consuming and labour--intensive. This study evaluated a novel method for automatic parsing of neurological assessment videos that is fast and reliable. 
Prior work that investigated facial movements in ALS used motion capture technology (e.g., electromagnetic articulography, optoelectronic techniques) or 3D scans of the face~\cite{Bandini01,bandini04,bandini05,quan2012facial}. The present work, using only 2D video data, represents a big step toward developing automated tools for assessing neurological disorders in home setting.

Overall, using the RepNet model let to a more consistent parsing of the BBP videos. Results show that RepNet achieved better parsing performance (IoU $> 65\%$ in all cases) and obtained parsing that separated HC and ALS participants based on repetition duration, with the difference between separations being significant (Table~\ref{table2}) (similar to manual parsing), as opposed to the landmark-based approach that mixed up the separation between ALS and HC groups (Table~\ref{table3} and Fig.~\ref{fig_BBP_reps}). This is because the movement in the BBP task is complex, consisting of multiple mouth opening/closings, and landmark-based parsing failed in detecting some of the repetitions. The complex movement also possibly caused a wider CI when the BBP task was parsed with RepNet; a potential solution is using audio-based or multi-modal (audio~+~video) parsing for more complex tasks that have an acoustic component, such as BBP. 

As seen in Table~\ref{table1}, both methods detected more variability (larger standard deviations) in ALS participants. This is in line with the nature of bulbar ALS, as previous studies have also shown more speech and pause variations across ALS groups~\cite{barnett}. Another observation was the decrease in IoU with repetition number (Fig.~\ref{fig_IOU}). A possible explanation for this could be that the RepNet model was trained on short videos ($\sim$10$\,$s), and parsing longer videos may have relatively lowered its parsing accuracy; a potential solution to this issue would be fine-tuning the model on longer clinical assessment videos.



In future work, we plan to extend this study to include a larger dataset with more speech and non-speech tasks, and also to expand the analysis to other clinical populations, e.g. PD and PS. In analyzing speech tasks, we plan to combine audio and video information, when appropriate, and evaluate RepNet performance when the TSM is generated from video, audio, or multi-modal information. An extended version of this approach can be used as part of an automated video-based assessment of bulbar ALS at home.

\section{CONCLUSION}
In this study, RepNet and a landmark-based method were evaluated in parsing video segments from ALS and HC groups performing tasks from common orofacial assessment examinations. The best results, in terms of IoU and also in terms of preserving the separation between ALS and HC, were obtained with RepNet.
Implementation of this video-based approach in clinical and home settings will help in monitoring disease progression and response to treatment. 
 early as possible.

\section*{Acknowledgment}

The authors would like to thank the Natural Sciences and Engineering Research Council of Canada (NSERC), ALS Canada, Parkinson Canada, Canadian Partnership for Stroke Recovery, 
Michael J. Fox Foundation and Weston Brain Institute, and the Kite Research Institute -- UHN for supporting our research. We would also like to thank Dr. Madhura Kulkarni for her role in supporting data collection and management for this project.



%
{\small
\bibliographystyle{IEEEtranBST/IEEEtran}
\bibliography{IEEEtranBST/IEEEfull}
}

\end{document}